\documentclass[conference]{IEEEtran}
\IEEEoverridecommandlockouts
\usepackage{cite}
\usepackage{amsmath,amssymb,amsfonts}
\usepackage{algorithmic}
\usepackage{graphicx}
\usepackage{textcomp}
\usepackage{xcolor}
\usepackage{stfloats} % set figure at current page
\usepackage{booktabs} % three rule
\DeclareUnicodeCharacter{2212}{-}
\def\BibTeX{{\rm B\kern-.05em{\sc i\kern-.025em b}\kern-.08em
    T\kern-.1667em\lower.7ex\hbox{E}\kern-.125emX}}
\begin{document}

\title{RTF-Q: Efficient Unsupervised Domain Adaptation with Retraining-free Quantization

\thanks{\textsuperscript{†} Corresponding author.}
% \thanks{We would like to thank Shenzhen Healall Technology Co., Ltd for sponsoring the research.}
}

\author{
\IEEEauthorblockN{Nanyang Du, Chen Tang, Yuxiao Jiang, Yuan Meng\textsuperscript{†}, Zhi Wang\textsuperscript{†}}
\IEEEauthorblockA{\textit{Shenzhen International Graduate School} \\
\textit{Tsinghua University}\\
Shenzhen, China \\
\{dny22, yx-jiang22\}@mails.tsinghua.edu.cn, 
tangchen18@outlook.com, \\yuanmeng@mail.tsinghua.edu.cn, wangzhi@sz.tsinghua.edu.cn}

% \and
% \and
% \and
% \and
% \and
}

\maketitle

\begin{abstract}
Performing unsupervised domain adaptation on resource-constrained edge devices is challenging. Existing research typically adopts architecture optimization (e.g., designing slimmable networks) but requires expensive training costs. Moreover, it does not consider the considerable precision redundancy of parameters and activations. 
To address these limitations, we propose efficient unsupervised domain adaptation with ReTraining-Free Quantization (RTF-Q). 
Our approach uses low-precision quantization architectures with varying computational costs, adapting to devices with dynamic computation budgets. 
We subtly configure subnet dimensions and leverage weight-sharing to optimize multiple architectures within a single set of weights, enabling the use of pre-trained models from open-source repositories. 
Additionally, we introduce multi-bitwidth joint training and the SandwichQ rule, both of which are effective in handling multiple quantization bit-widths across subnets.
Experimental results demonstrate that our network achieves competitive accuracy with state-of-the-art methods across three benchmarks while significantly reducing memory and computational costs.

% Code is available at \href{https://github.com/dunanyang/RTF-Q}{https://github.com/dunanyang/RTF-Q}.
\end{abstract}

\section{Introduction} 
Unsupervised Domain Adaptation (UDA)~\cite{REDA,UDA_survey,UDA_review_2020} seeks to align source and target domains in a shared representation space, mitigating domain shifts between training data and edge device environments to improve generalization. \cite{NAS_DA} attempts to learn optimal architectures to further boost the performance on the target domain with NAS~\cite{NAS,BigNAS}. SlimDA~\cite{SlimDA} enhances cross-domain generalization by integrating a weight-sharing module into SymNet~\cite{SymNet} and sampling subnets from the model bank with the Sandwich rule for ensemble learning. Recently, AnyDA~\cite{AnyDA} introduced a subnet architecture with three configurable dimensions—depth, width, and resolution—allowing network slicing to tailor subnets for different edge devices. Additionally, AnyDA employs a self-supervised learning~\cite{SS-Self_Distillation} with knowledge distillation~\cite{Knowledge_Distilling} called bootstrapped recursive distillation approach and information maximization~\cite{IM-Loss} of target data as a regularizer. However, these methods rely on full-precision networks, ignoring numerical precision redundancy, which increases computational and memory demands—unsuitable for resource-limited edge devices~\cite{Int-Q,Int4}. Furthermore, many approaches require training all subnets from scratch, effectively doubling the computational cost due to the added pre-training stage.

In this paper, we propose an efficient unsupervised domain adaptation method with retraining-free quantization, the first to leverage quantization for UDA, optimizing it for resource-constrained edge devices. First, we apply low-precision quantization to UDA, improving efficiency and adaptability to devices with dynamic computation budgets. Additionally, we configure subnet dimensions with minimal performance impact and employ weight-sharing to optimize multiple architectures using a single set of weights, enabling the use of pre-trained models and significantly reducing training costs. Finally, our SandwichQ method trains multiple quantization bit-widths simultaneously, improving performance over single-bit quantization approaches.
Specifically, our subnets are defined by three dimensions: width, input resolution, and quantization bit-width. We employ a teacher-student model for knowledge distillation to balance subnet performance. To address the lack of labeled data in the target domain, we generate pseudo-labels via the student supernet, using them as supervisory signals to ensure a discriminative latent space. Subnets are tested under varying computational budgets to optimize performance for different edge devices, enhancing the practicality and efficiency of UDA. Our key contributions are as follows:

\begin{itemize}
\item We are the first work to consider efficient UDA from the perspective of low-precision quantization and design a retraining-free quantization scheme to further enhance the efficiency of UDA.

\item By subtly configuring subnet dimensions, we eliminate the need for pre-training from scratch on ImageNet-1K, reducing the training stage and making the training more economical.

\item We propose joint training with multiple quantization bit-widths and the SandwichQ rule for the joint training across various quantization bit-widths, which resulted in 5.6 and 1.4 improvements in classification accuracy, respectively, with improved training efficiency.
\end{itemize}

\section{Related Work}
\subsection{Retraining-Free Domain Adaptation}
Retraining-free domain adaptation \cite{AnyDA,SlimDA} is a technique that applies retraining-free methods \cite{S-Net,US-Net,AutoSlim} to UDA tasks, enabling subnets to meet different computation budget requirements of edge devices. However, due to significant structural configuration differences between subnets and the supernet \cite{US-Net,DSBN}, they all require pre-training from scratch on ImageNet-1K. In addition, none of them consider the computational and memory pressure caused by parameter precision redundancy \cite{Q_Survey,Adabits} on edge devices. This paper addresses these issues through reasonable structural configurations and network quantization\cite{LSQ,ABN}.

\subsection{Efficient Domain Adaptation}
Although traditional UDA methods have developed rapidly, efficient UDA still has many aspects that need to be studied. To accelerate UDA inference, MSDNet \cite{MSDNet} uses the DANN \cite{DANN} method and proposes an early-exit architecture. \cite{REDA,DDA} further proposes a multi-scale mechanism. The authors of~\cite{TCP,small_is_beautiful,CPC} use network pruning methods. Network quantization is one of the most promising compression methods for deploying deep neural networks on edge devices \cite{ABN}. This paper uses quantization-aware training (QAT) \cite{LSQ} to compress and accelerate the retraining-free UDA network while ensuring classification accuracy.

\subsection{Network Quantization}
Network quantization~\cite{Q_Survey,DeepCompression,Compression-Through-Dis-and-Quant,TC-MP_Layer} has been extensively researched to reduce memory usage and computational complexity by representing weights and activations with lower digital bit-widths, which will reduce computational complexity and memory usage \cite{Lsq+,PACT,Adabits}, enabling more efficient deployment on resource-constrained devices without significantly sacrificing accuracy~\cite{Int4,Int-Q}. \cite{TC-RTF} devise a one-shot training searching paradigm for mixed-precision model compression. Recent quantization methods, such as LSQ \cite{LSQ}, have achieved good results because they make the quantization step size a learnable parameter. 
\section{METHOD}

\subsection{Preliminaries}
Formally, we have two domains with different data distributions but a common label space $L$. Among them, $X_s$ is the labeled source domain, and $X_t$ is the unlabeled target domain, with sizes of $N_s$ and $N_t$, respectively. We also have $n$ ascending sorted computational budgets $\mathcal{B} = \{b_1, b_2, \dots, b_n\}$ for inference. Additionally, our quantization bit-width search space is $\mathcal{Q} = \{q_1, q_2, \dots, q_m\}$.

\subsection{Approach Overview}
\begin{figure*}[ht]
    \centering
    \includegraphics[width=0.9\linewidth]{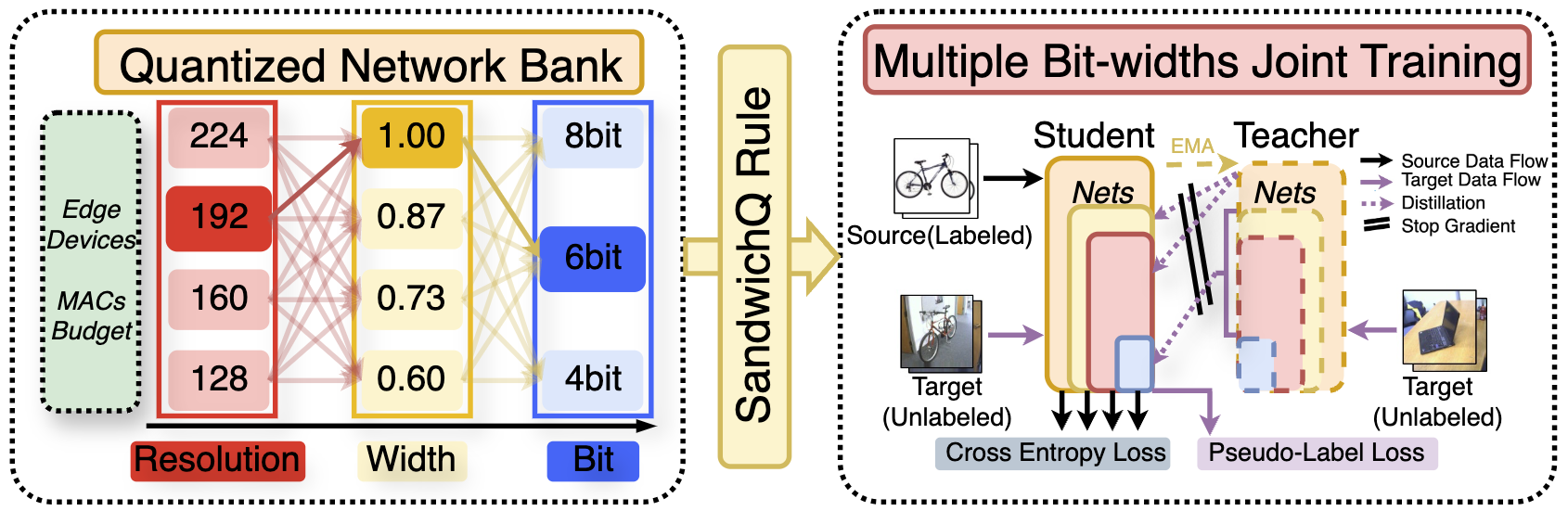}
    \caption{The overall framework of RTF-Q. Left: We first configure subnets employing weight-sharing and low-precision quantization architectures with varying MACs' budgets, forming a quantized network bank. Right: 
    We employ the proposed SandwichQ rule to sample a fixed number of networks and optimize their shared parameters using joint training of multiple bit-widths. Only the student's corresponding subnet that meets the MAC budget is used during inference.}
    \label{fig:framework}
\end{figure*}

Fig. \ref{fig:framework} provides an overview of our method. Firstly, we use quantization dimensions to provide a more lightweight network for edge devices. Through weight sharing, we obtained a total of 48 subnets with varying computation budgets, allowing dynamic adaptation to the computational needs of edge devices under different budget constraints. Secondly, we obtained the optimal configuration dimensions for subnets: resolution, width, and quantization bit-width. In this configuration, the initialization of subnets directly uses open-source weight files without the need for expensive pre-training. We optimize the common parameters of multiple subnets through joint training with multiple-bit widths and the SandwichQ rule. During inference, we will use only the student network. Given any computation budget $b_i \in \mathcal{B}$, we obtain all possible configuration tuples within the budget $b_i$ from the $\mathcal{R} \times \mathcal{W} \times \mathcal{Q}$ subnets. We select the best-performing subnet and report its classification accuracy. Additionally, we adopt a teacher-student model composed of two switchable quantized networks with the same architecture, parameterized by $\theta_{stu}$ and $\theta_{tea}$, respectively, forming $\mathcal{Q}^{\theta_{stu}}$ and $\mathcal{Q}^{\theta_{tea}}$. Each has $n$ subnets composed of partial parameters of the supernet. Among them, $\mathcal{Q}^{\theta_{stu}}$ is updated by training with QAT \cite{LSQ}, and $\mathcal{Q}^{\theta_{tea}}$ is updated by the exponential moving average (EMA) of the $\mathcal{Q}^{\theta_{stu}}$. 

The student network is trained using supervised learning on the source domain. The teacher network only inputs target domain data and provides cross-domain knowledge for domain alignment to the student network under lower computation budgets. Additionally, we use self-supervision to utilize classification information, ensuring the discriminative latent space of the unlabeled target domain through pseudo-label loss on the student supernet. We also use information maximization of target data as a regularizer.

\subsection{Quantization-Aware Training}
We aim to quantize networks under various computational budgets to minimize the loss in classification accuracy. We use LSQ, where the quantization step size is learnable for QAT. Specifically, we use low-precision integer operations to compute convolutional layers, requiring quantization of weights and activations. Given full precision data to quantize $v$, a quantizer with step size $s$, and quantization bounds $Q_P$ and $Q_N$, the network is quantized. 
We define a quantizer that computes $\bar{v}$, a quantized and integer scaled representation of the data, and $\hat{v}$, a quantized representation of the data at the same scale as $v$.

\begin{equation}
    \bar{v} = \lfloor clamp(v/s, -Q_{N}, Q_{P}) \rceil
\end{equation}
\begin{equation}
    \hat{v} = \bar{v} \times s
\end{equation}

where \textit{clamp($z, r_1, r_2$)} returns $z$, clamping values lower than $r_1$ to $r_1$ and higher than $r_2$ to $r_2$. $\lfloor z \rceil$ returns the nearest integer. Given a $b$-bit encoding, for unsigned data (e.g., activation values), $Q_N = 0$ and $Q_P = 2^b−1$, for signed data (e.g., convolution weights), $Q_N = −2^{b−1}$ and $Q_P = 2^{b−1}−1$.

\subsection{Subnet Dimensions}
We typically have several configurable dimensions to obtain subnets with different computation budgets: depth (number of layers), width, input resolution, and kernel size. The subnet is composed of partial parameters of the supernet. For retraining-free UDA tasks, we need domain-specific batch normalization layers and switchable modules that are very different from official ResNet \cite{ResNet}, complicating network initialization. Previous methods required pretraining all subnets on ImageNet-1K from scratch, a dataset significantly larger than all task datasets combined. For each new configuration, pretraining is required, which is extremely costly. Furthermore, as stated in S-Nets \cite{S-Net,Mobilenetv2}, reducing depth compared to reducing width does not reduce inference memory usage on edge devices.

\subsection{Joint Training}
We can perform quantization with specific bit-widths to obtain networks with specific quantization bit-widths. Generally, higher bit-width networks perform better than lower bit-width ones, necessitating the use of high bits for quantization. To address this, we use a method similar to ensemble learning, jointly learning across multiple quantization bit-width spaces. This allows lower bit-width networks to learn corresponding knowledge from higher bit-width networks, achieving better performance. 

\subsection{SandwichQ Rule}
For networks with multiple quantization bit-widths, especially when the number of subnets doubles, determining the number of training subnets and training strategies becomes crucial. Standard Sandwich \cite{US-Net} training involves the largest, smallest, and limited number of intermediate subnets. 
However, an independent network is required for each quantization bit-width. Therefore, the Sandwich rule must be applied to the network under each bit-width quantization. We found that standard Sandwich incurs high training costs and yields suboptimal network results, possibly caused by the coupling or overfitting of multiple quantization bit-widths. Hence, we propose the SandwichQ rule.
Specifically, in each time, we train students on a minnet with the lowest quantization bit-width, a maxnet with the highest quantization bit-width, and a limited number of subnets with randomly selected quantization bit-width among exponentially increasing subnets. Under the configuration of this paper, it is only necessary to train 4 subnets out of the 48 networks.

\section{Experiments}
\subsection{Experimental Setting}

\textbf{Datasets and Tasks.} We use 3 benchmarks to evaluate the performance of the proposed method: Office-31 \cite{office31}, Office-Home \cite{Office-Home}, and DomainNet \cite{DomainNet}. Office-31 contains a total of 4,652 images from 31 categories across 3 different domains. Office-Home is a challenging dataset with a total of 15,500 images from 4 different domains containing 65 categories. DomainNet is the largest available benchmark dataset, containing images from 6 domains, with 600K images distributed across 345 categories. There are 48 unsupervised domain adaptation tasks in total.

\textbf{Baselines.} We compare our approach with the following baselines: ResNet \cite{ResNet}, MobileNetV3 \cite{MobileNetV3}, MSDNet \cite{MSDNet}, REDA\cite{REDA} equipped with DANN \cite{REDA}, TCP \cite{TCP} and DDA \cite{DDA}, as well as the SOTA work AnyDA \cite{AnyDA}.

\textbf{Implementation Details.} We use the ResNet-50 architecture as the supernet. For budget configurations, we use $R = \{224, 192, 160, 128\}, W = \{1.00, 0.86, 0.73, 0.60\}, Q = \{8, 6, 4\}$, providing us with 48 computation budgets ranging from approximately $0.2\times 10^9$ to $2.0\times 10^9$ MACs for subnets, while reducing BitOPs \cite{BOPs,Q-Diffusion} by about 16 times. We further divide the budget range into 8 equal intervals. For all experiments, we use a batch size of 256 (128 sources + 128 targets) per GPU. We use a learning rate of $2e^{-4}$ for the Office-31 and Office-Home datasets and $3e^{-5}$ for the DomainNet dataset. We update the learning rate using cosine annealing. We report the average classification accuracy.

\subsection{Results and Analysis}
\begin{figure*}[ht]
    \centering
    \includegraphics[width=0.85\linewidth]{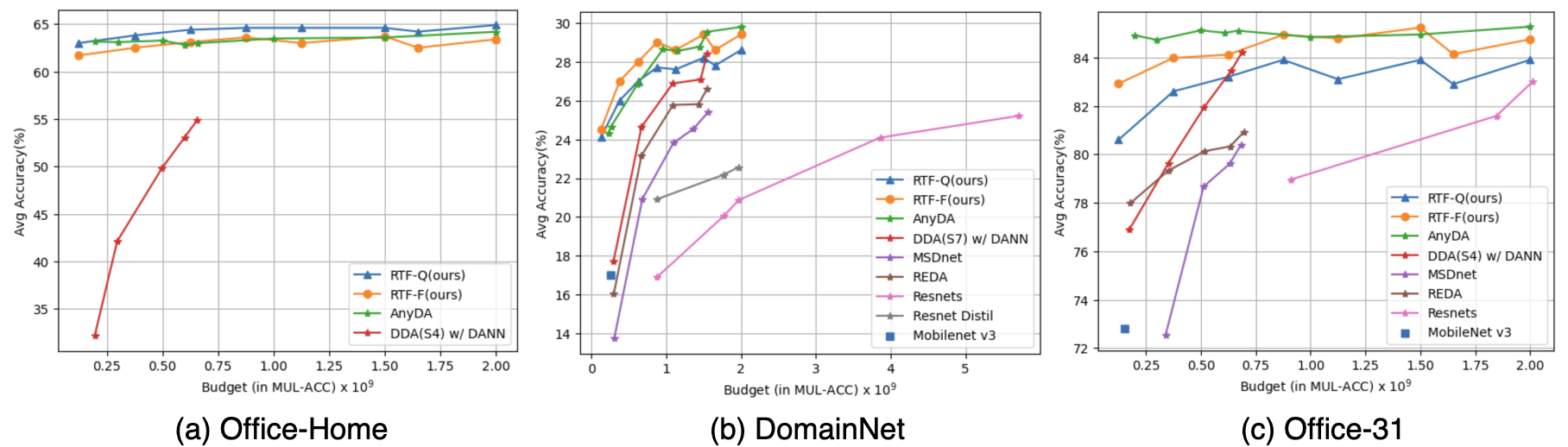}
    \caption{The average classification accuracy of RTF-Q and RTF-F on the UDA tasks of Office-31, DomainNet, and Office-Home.}
    \label{fig:full_q}
\end{figure*}

Fig. \ref{fig:full_q} shows the experimental results of RTF-Q on the Office-31, DomainNet, and Office-Home datasets. RTF-Q outperforms all methods except AnyDA in terms of accuracy across all computation budget intervals. It exceeds AnyDA on Office-Home and maintains comparable accuracy on DomainNet and Office-31. The results indicate that our RTF-Q can still perform well in UDA tasks without retraining, even with less than one-sixteenth of BitOPs used. Additionally, we provide the full precision 32-bit result of RTF-Q, which is RTF-F. The classification accuracy of RTF-F and AnyDA is comparable. It is worth noting that the results are achieved without pretraining on ImageNet-1K from scratch, demonstrating that our selected subnet dimensions are well-suited for initialization with open-source weight files.

\subsection{Ablation Studies}
\textbf{Impact of Depth on Subnets.} As the average classification accuracy of experiments on Office-31 and Office-Home is shown in Table \ref{tab:depth_impact}, without pre-training ImageNet-1K from scratch, the depth dimension will decrease the classification ability of the subnet by 5 percentage points. When the depth dimension of the subnet is removed, our RTF-F, which does not require pre-training for ImageNet-1K from scratch, achieves results comparable to AnyDA. This more optimal subnet configuration dimension allows us to save significant pre-training costs.

\begin{table}[!ht]
    \centering
    \caption{The Impact of Depth Dimension on Classification Accuracy.}
    \label{tab:depth_impact}
    \begin{tabular}{ccccc}
        \toprule
        ~     & \multicolumn{2}{c}{Office-31} & \multicolumn{2}{c}{Office-Home}                        \\
        ~     & w/ depth                     & w/o depth                      & w/ depth & w/o depth \\ \midrule
        AnyDA & 85.1                         & -                              & 63.6     & -         \\
        Ours(RTF-F)  & 80.4                         & 84.4                           & 57.8     & 62.9      \\ \bottomrule
    \end{tabular}
\end{table}

\begin{figure}[ht]
    \centering
    \includegraphics[width=0.7\linewidth]{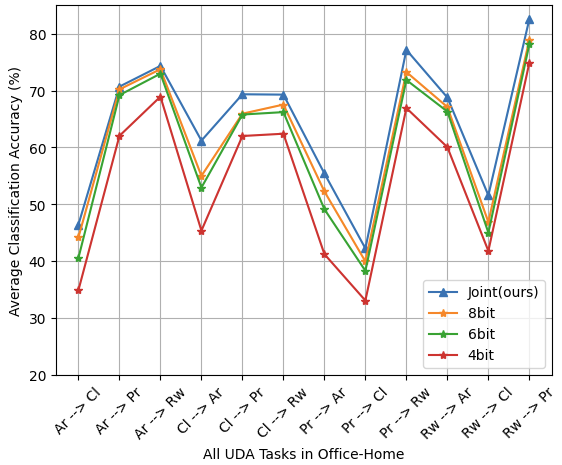}
    \caption{Effect of joint training across multiple quantization bit-widths on classification accuracy.}
    \label{fig:multi_bit}
\end{figure}

%\noindent
\textbf{Joint Training of Multiple Bit-widths.} We compare two training methods: training with only a single quantization bit-width and training with multiple quantization bit-widths. The classification accuracy of 12 tasks in Office-Home is shown in Fig. \ref{fig:multi_bit}, our joint learning method with multiple quantization bit-widths achieves higher accuracy than training with a single bit-width alone, indicating that multiple bit-widths can learn from each other, thereby improving the classification accuracy of a single bit-width network.

% [64.08958333 61.25416667 59.70416667 54.47291667] 5.6125000000000185

%\noindent
\textbf{SandwichQ Rule.} 
We refer to the method of sampling the maxnet, two random subnets, and the minnet each time as SandwichQ. As shown in the Table \ref{tab:SandwichQ}, using the standard Sandwich or a larger number of intermediate sub-networks results in suboptimal performance. The results show that SandwichQ is more conducive to model convergence and has better results than Sandwich.
\begin{table}[!ht]
    \centering
    \caption{Results on Office-Home Classification with Different Sampling Rules During Training.}
    \label{tab:SandwichQ}
    \begin{tabular}{cccc}
        \toprule
        Rule & max+6 random+min & Sandwich & SandwichQ \\
        \midrule
        Acc &   62.0 & 63.7 & \textbf{64.3}   \\ 
        \bottomrule
    \end{tabular}
\end{table}

\section{Conclusion}
In this paper, we propose efficient unsupervised domain adaptation with retraining-free quantization. We reduce the BitOPs at least 16 times compared to the baseline, and meet the requirements of dynamic network computation budgets for edge devices. 
By subtly configuring subnet dimensions, we eliminate the need for pretraining from scratch on ImageNet. 
Joint training with multiple quantization bit widths and the SandwichQ rule achieve better classification results. 
Extensive experiments are carried out to validate the effectiveness of RTF-Q. In short, our work provides a practical UDA framework for real-world scenarios, and we hope it can inspire new ideas to make UDA more efficient.

% \section*{Acknowledgment}

% The preferred spelling of the word ``acknowledgment'' in America is without 
% an ``e'' after the ``g''. Avoid the stilted expression ``one of us (R. B. 
% G.) thanks $\ldots$''. Instead, try ``R. B. G. thanks$\ldots$''. Put sponsor 
% acknowledgments in the unnumbered footnote on the first page.

% \section*{References}

\bibliographystyle{IEEEtran}
\bibliography{IEEEtranBST2/IEEE_ref}

\end{document}